# D-FINE-SEG: OBJECT DETECTION AND INSTANCE SEGMENTATION FRAMEWORK WITH MULTI-BACKEND DEPLOYMENT

Argo Saakyan, Dmitry Solntsev

Veryfi Inc.

## ABSTRACT

Transformer-based real-time object detectors achieve strong accuracy-latency trade-offs, and D-FINE is among the top-performing recent architectures. However, real-time instance segmentation with transformers is still less common. We present D-FINE-seg, an instance segmentation extension of D-FINE that adds: a lightweight mask head, segmentation-aware training, including box cropped BCE and dice mask losses, auxiliary and denoising mask supervision, and adapted Hungarian matching cost. On the TACO dataset, D-FINE-seg improves F1-score over Ultralytics YOLO26 under a unified TensorRT FP16 end-to-end benchmarking protocol, while maintaining competitive latency. Second contribution is an end-to-end pipeline for training, exporting, and optimized inference across ONNX, TensorRT, OpenVINO for both object detection and instance segmentation tasks. This framework is released as open-source under the Apache-2.0 license. GitHub repository - https://github.com/ArgoHA/D-FINE-seg.

***Index Terms*** — Computer Vision, Object Detection, Instance Segmentation, Deployment

## 1. INTRODUCTION

Computer Vision often shines in real-time settings and instance segmentation is fundamental in the field. DETR [1] family detectors are elegant because of their end-to-end pipeline - decoder allows to directly output bounding boxes 1 to 1 without the need for non-max suppression (NMS) which is an advantage for real-time inference. However, instance segmentation in practice often forces heavier heads and latency starts becoming a problem.

We show that D-FINE [2] can be extended with a lightweight mask head while achieving competitive accuracy, keeping both low latency and exportability to popular formats for optimal inference on server hardware and edge devices.

Contributions:
- Lightweight mask head design for D-FINE encoder outputs;
- Segmentation-aware training, loss/matcher + aux/denoising supervision;
- Reproducible multi-backend deployment protocol.

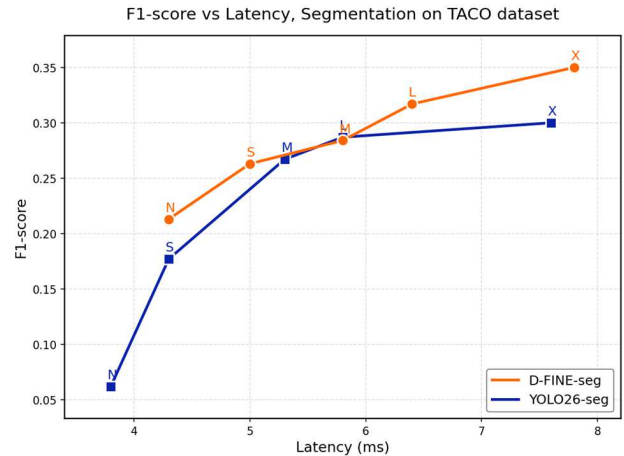

*Figure 1. End-to-end accuracy and latency benchmarking, segmentation task*

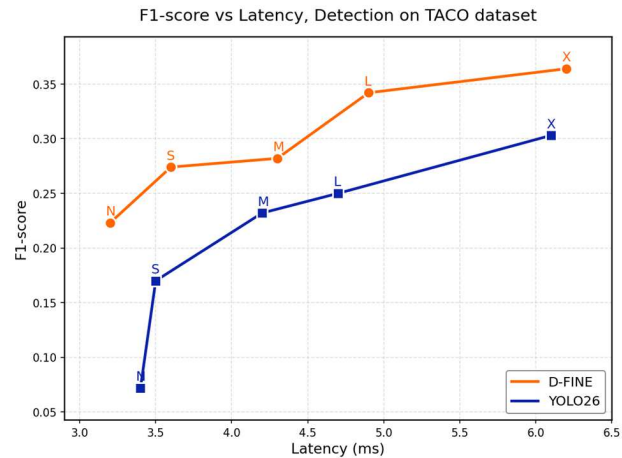

*Figure 2. End-to-end accuracy and latency benchmarking, detection task*

## 2. RELATED WORK

The heart of D-FINE-seg is the original D-FINE architecture which itself was based on RT-DETR [3]. Mask head was inspired by Mask DINO [4]. Training pipeline is custom, but still inspired by projects like: DAMO-YOLO [5], YOLOX [6], Ultralytics YOLO [7] and YOLOv4 [8].

Mask2Former [9] unified segmentation tasks with masked attention, while Mask DINO extended DINO [10] with a mask prediction branch using dot-product between query embeddings and pixel features. SAM [11] demonstrated the power of foundation models for segmentation but targets a different use case (promptable segmentation). Real-time transformer-based instance segmentation remains less explored compared to detection.

Finally - exporting, optimizations and deployment rely on ONNX [12], TensorRT [13] and OpenVINO [14].

## 3. METHOD

Object detection part inherits the original D-FINE with two main ideas - Fine-grained Distribution Refinement (FDR) and Global Optimal Localization Self-Distillation (GO-LSD). FDR iteratively refines probability distributions instead of predicting fixed bounding box coordinates which provides better localization accuracy. GO-LSD passes knowledge from the final Decoder layer to earlier layers, playing a role of a self-distillation.

The architecture consists of the Backbone (CNN) for feature extraction, HybridEncoder for multi-scale feature fusion (FPN + PAN), Transformer Decoder with contrastive denoising accelerating convergence and improving matching quality. The model comes in multiple sizes (N/S/M/L/X) with varying backbone depths and decoder widths.

Our mask head design follows Mask DINO's paradigm - projecting decoder queries to mask embeddings and computing masks via dot product with pixel features - but unlike Mask DINO, which incorporates stride-4 backbone features for high-resolution detail, our mask head operates solely on HybridEncoder's PAN outputs (stride 8/16/32) and uses a single transposed convolution (stride-2) to reach 1/4 resolution. Leveraging the encoder's already fused multi-scale representations, we avoid passing stride-4 backbone features to the mask head.

### 3.1. Architecture

Ground truth (GT) segmentation masks are preprocessed to a format [K, H, W], where K is the number of objects in the image. Each such mask is binary. That's also the output format of the model.

An overview of model architecture is shown on Figure 3. Mask head architecture – Figure 4.

Mask head receives feature maps produced by HybridEncoder (from multi-scale PAN). Mask head consists of:
1. Per-level (feature stride 8/16/32) 1x1 projection + GroupNorm to a common channel dimension (default is 256).
2. Fusion into stride-8 resolution (bilinear upsample to the finest feature stride and summation).
3. 3×3 convolution + GroupNorm + ReLU for mask smoothness.
4. Bilinear upsampling followed by 3×3 convolution + GroupNorm + ReLU to reach 1/4 scale (instead of transposed convolutions).

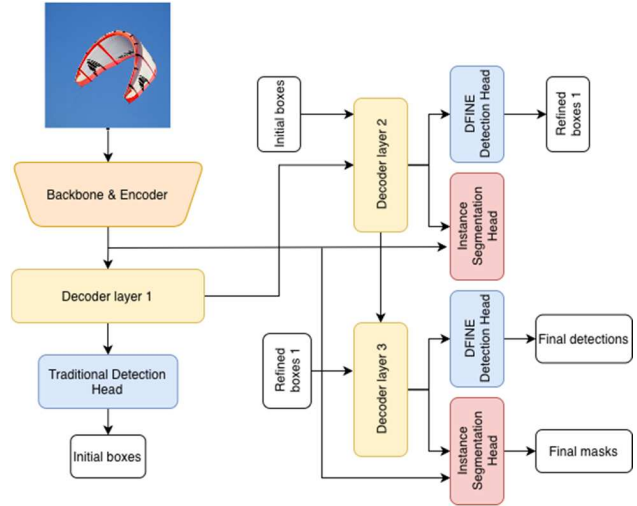

*Figure 3. D-FINE-seg model architecture*

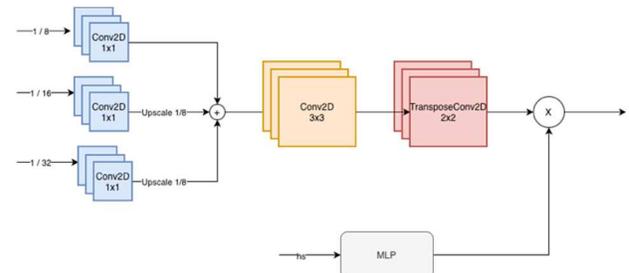

*Figure 4. D-FINE-seg mask head architecture*

Next step is projecting each query to a per-instance mask embedding, passing per-layer decoder hidden states to a 3-layer MLP. Mask logits are then computed as a scaled (for stabilization of magnitudes) dot product between per-query mask embeddings and the shared per-image mask feature map, which is equivalent to a dynamic 1x1 convolution over channels. Logits are produced at H/4 × W/4 and are later resized to the original size during postprocessing.

### 3.2. Auxiliary and denoising mask supervision

When training a segmentation task, mask logits are computed not only for the last Decoder layer, but also for intermediate layers (auxiliary outputs). Denoising masks are computed for denoising queries and supervised using the same cropped mask loss. This type of supervision helps improve accuracy without affecting inference latency or memory usage (only adds time and memory usage during training).

### 3.3 Loss functions

Originally criterion in D-FINE consisted of:
- Classification: Varifocal loss (VFL) [15] and optional Focal loss [16];
- Box regression: L1, Generalized Intersection over Union (GIoU) [17];
- D-FINE specific losses: Fine-Grained Localization (FGL), Decoupled Distillation Focal (DDF);

For the segmentation task we are adding:
- Box cropped mask binary cross entropy (BCE) computed inside the region of interest (ROI) and normalized by ROI area (mean loss per ROI pixel);
- Box cropped mask dice loss, computed on sigmoid probabilities.

Mask losses are computed only inside the matched GT box. Both losses are averaged over matched instances. GT instance masks are resized to the mask head resolution using bilinear interpolation, producing soft targets. Following D-FINE, we apply the full loss suite to the final decoder layer and to intermediate decoder layers as auxiliary losses. Loss weights used: VFL: 1, L1: 5, GIoU: 2, FGL: 0.15, DDF: 1.5, mask BCE: 1, mask dice: 1.

### 3.4. Hungarian matcher

Hungarian matcher is used to assign one predicted object (query) with precisely one GT object. Matching cost is a weighted sum of classification, L1 (box distance) and GIoU. In our implementation we are adding:
- Dice overlap cost (1 - Dice) between predicted mask probabilities and resized GT masks;
- Sigmoid focal mask cost between predicted mask logits and resized GT masks (averaged over mask pixels);
- Mask costs are full-map at the mask-head output resolution (unlike ROI-cropped losses).

### 3.5. Postprocessing

1. Filter instances when confidence score is lower than a threshold.
2. Mask resize from 1/4 scale to the original image size via bilinear interpolation.
3. Mask binarization based on a threshold.
4. Cleanup via zeroing mask pixels outside the corresponding bounding box. This is consistent with our training objective, which emphasizes supervision inside object ROIs.

## 4. IMPLEMENTATION

The D-FINE-seg framework provides capabilities for training (on custom datasets), benchmarking, exporting, and running inference with object detection and instance segmentation models. While we adopted the original D-FINE architecture, losses, and Hungarian matcher, everything else was reimplemented from scratch with an added segmentation head. This natively resolved issues that original pipeline had with training on some custom datasets [18].

Currently, pretrained weights for the mask head are not available in D-FINE-seg. Consequently, we fine-tune the model by initializing the mask head from scratch while loading COCO-pretrained weights for the backbone and the remaining detection components. Pretraining the mask head on COCO remains an important direction for future work and may improve mask boundary precision.

### 4.1. Key features:
1. Object detection and instance segmentation tasks fully supported.
2. Single config file for all settings.
3. Grouped learning rates for backbone vs decoder.
4. Mosaic and other augmentations.
5. Exponential moving average (EMA) model evaluation and checkpointing.
6. Gradient accumulation.
7. DDP support for multi-GPU training.
8. Mask-aware validation and practical detection metrics (F1-score, Precision, Recall).
9. Memory-efficient mask handling with run-length encoding (RLE) and batched evaluation.
10. WandB and local logging.
11. Benchmarking across formats, batch sizes, and quantization levels.
12. Error analysis script to identify potential annotation mistakes.

### 4.2. Export and inference capabilities:
1. Multi-format export (ONNX, TensorRT, OpenVINO).
2. FP16 support.
3. INT8 accuracy-aware quantization (OpenVINO);
4. Format-specific optimized inference code.

### 4.3. Typical workflow:
1. Configure dataset path and parameters in the config file.
2. `make train` - train the model.
3. `make export` - export to supported formats.
4. `make infer` - for inference or "infer" modules for the final deployment.

## 5. EXPERIMENTS

We evaluated D-FINE-seg against Ultralytics' latest YOLO26 in a fine-tuning scenario, initializing both models with COCO-pretrained weights. We are trying to be as close to real-world as possible:
1. Fixed-threshold metrics. We report F1-score, Precision, Recall, and IoU at a fixed confidence threshold, matching how models are actually deployed.
2. Unified accuracy-latency measurement. Models are converted to TensorRT FP16, then both accuracy and latency are measured on that model simultaneously.
3. End-to-end latency. Includes preprocessing, forward pass, postprocessing. Reading an image from the disk is excluded. Raw inference latency (forward pass) is also included as a separate value.

Throughout the paper, we refer to the segmentation model as D-FINE-seg and detection as D-FINE, although both were trained with our D-FINE-seg framework.

## 5.1. Metric Calculation
1. One-to-one matching: each GT object matches to only one prediction.
2. A match counts as a TP if IoU (mask for segmentation and box for detection) > 0.5 and class IDs match.
3. If multiple predictions overlap the same GT object, only the highest IoU match is counted as TP, remaining ones are FPs.
4. Class IDs must also match. If GT class and prediction class do not match – it's one FP and one FN.
5. IoU-score (penalized): we report the mean IoU over the evaluation outcomes defined above, where class-correct TPs contribute their IoU and all FPs and FNs contribute 0 (i.e., averaged over TPs + FPs + FNs).

To measure end-to-end latency, we warmed up each model on 10 samples and then followed these steps for each image from validation set (212 images):
1. Load the image using OpenCV (outside of the latency calculation).
2. Synchronize the GPU and start the timer.
3. Run inference. Both models perform preprocessing (resize, normalization, GPU transfer), forward pass, and postprocessing (box scaling, mask resize to the original image size and cleanup) internally, and return labels, confidence scores, bounding boxes, and full-resolution masks.
4. Synchronize the GPU and stop the timer.

Details on how to reproduce our results can be found in the D-FINE-seg's repository in `paper_assets` folder.

## 5.2. Dataset
TACO [19] - 1,500 images, 60 categories of *waste* in diverse environments. One category has no instances, so we report results over 59 effective classes. Split 86/14 (train/validation) by batch ID to prevent data leakage. We trained both segmentation (polygon annotations) and detection (bounding box annotations) variants using YOLO style annotations. Metrics were calculated on the validation set.

## 5.3. Configuration
Both models were trained with 640×640 input size. YOLO26 was trained for 100 epochs and D-FINE-seg for 50 as it converges faster. Both models exported to TensorRT with FP16 precision.

We used each framework's default confidence threshold (0.25 for YOLO26 and 0.5 for D-FINE-seg). Because confidence score distributions differ between those models, using each framework's default threshold provides a reasonable per-model operating point for precision–recall trade-offs.

Hardware: NVIDIA RTX 5070 Ti (16GB) and Intel i5 12400f. Software: CUDA Version: 12.8. TensorRT Version: 10.10.0.31. Ultralytics version: 8.4.6. Batch size 1.

*Table 1. COCO style mask APs*

| Model | Mask mAP@50-95 | Mask mAP@50 |
|---|---|---|
| **D-FINE-seg N** | **0.094** | **0.141** |
| YOLO26-seg N | 0.041 | 0.058 |
| **D-FINE-seg S** | **0.177** | **0.250** |
| YOLO26-seg S | 0.111 | 0.165 |
| D-FINE-seg M | 0.157 | 0.229 |
| YOLO26-seg M | **0.195** | **0.270** |
| **D-FINE-seg L** | **0.212** | **0.310** |
| YOLO26-seg L | 0.174 | 0.242 |
| **D-FINE-seg X** | **0.242** | **0.340** |
| YOLO26-seg X | 0.210 | 0.291 |

*Table 2. COCO style detection APs*

| Model | Box mAP@50-95 | Box mAP@50 |
|---|---|---|
| **D-FINE N** | **0.123** | **0.169** |
| YOLO26 N | 0.060 | 0.075 |
| **D-FINE S** | **0.202** | **0.244** |
| YOLO26 S | 0.098 | 0.124 |
| **D-FINE M** | **0.204** | **0.246** |
| YOLO26 M | 0.172 | 0.214 |
| **D-FINE L** | **0.256** | **0.314** |
| YOLO26 L | 0.230 | 0.272 |
| **D-FINE X** | **0.269** | **0.336** |
| YOLO26 X | 0.256 | 0.300 |

*Table 3. Format comparisons*

| Model | Format | F1-score | Latency (ms) |
|---|---|---|---|
| D-FINE-seg S | Torch FP32 | 0.263 | 20.4 |
| D-FINE-seg S | TensorRT FP32 | 0.264 | 6.5 |
| D-FINE-seg S | TensorRT FP16 | 0.263 | 5.0 |
| D-FINE S | Torch FP32 | 0.276 | 18.0 |
| D-FINE S | TensorRT FP32 | 0.272 | 4.5 |
| D-FINE S | TensorRT FP16 | 0.274 | 3.6 |

## 5.4. Results
Main metrics are presented in the Table 4, Figures 1 and 2. In segmentation task we observe a mean relative F1-score improvement of **~65%** across N/S/M/L/X over YOLO26 under our TACO fine-tuning setup, with ~10% mean relative latency overhead. In detection task - ~**70%** higher F1-score and ~1% latency overhead.

We also measured COCO style metrics (Tables 1 and 2), setting a low threshold (0.01) to retain predictions, AP was computed in the standard COCO manner for both segmentation and detection models, with the maximum number of detections per image set to 100.

*Table 4. Comparison of object detection and instance segmentation tasks with YOLO26.*

| Model | #Params (M) | F1-score | IoU | Precision | Recall | Latency (ms) | Raw inference latency (ms) |
|---|---|---|---|---|---|---|---|
| *Segmentation task* | | | | | | | |
| D-FINE-seg N | 5.1 | **0.213** | 0.095 | 0.272 | 0.175 | 4.3 | 1.4 |
| YOLO26-seg N | 2.7 | 0.062 | 0.027 | 0.272 | 0.035 | 3.8 | 1.1 |
| D-FINE-seg S | 11.9 | **0.263** | 0.125 | 0.339 | 0.215 | 5.0 | 2.2 |
| YOLO26-seg S | 10.4 | 0.177 | 0.080 | 0.278 | 0.130 | 4.3 | 1.4 |
| D-FINE-seg M | 21.2 | **0.284** | 0.134 | 0.316 | 0.258 | 5.8 | 2.8 |
| YOLO26-seg M | 23.6 | 0.267 | 0.128 | 0.365 | 0.210 | 5.3 | 2.1 |
| D-FINE-seg L | 32.8 | **0.317** | 0.152 | 0.369 | 0.278 | 6.4 | 3.5 |
| YOLO26-seg L | 28.0 | 0.287 | 0.137 | 0.394 | 0.226 | 5.8 | 2.7 |
| D-FINE-seg X | 64.3 | **0.350** | 0.172 | 0.391 | 0.318 | 7.8 | 4.8 |
| YOLO26-seg X | 62.8 | 0.300 | 0.146 | 0.408 | 0.238 | 7.6 | 4.3 |
| *Detection task* | | | | | | | |
| D-FINE N | 3.8 | **0.223** | 0.108 | 0.295 | 0.180 | 3.2 | 1.2 |
| YOLO26 N | 2.4 | 0.072 | 0.033 | 0.274 | 0.042 | 3.4 | 1.0 |
| D-FINE S | 10.3 | **0.274** | 0.140 | 0.327 | 0.240 | 3.6 | 1.6 |
| YOLO26 S | 9.5 | 0.170 | 0.081 | 0.279 | 0.122 | 3.5 | 1.2 |
| D-FINE M | 19.6 | **0.282** | 0.147 | 0.342 | 0.239 | 4.3 | 2.3 |
| YOLO26 M | 20.4 | 0.232 | 0.115 | 0.303 | 0.188 | 4.2 | 1.7 |
| D-FINE L | 31.2 | **0.342** | 0.180 | 0.409 | 0.294 | 4.9 | 2.9 |
| YOLO26 L | 24.8 | 0.250 | 0.128 | 0.356 | 0.193 | 4.7 | 2.3 |
| D-FINE X | 62.6 | **0.364** | 0.195 | 0.394 | 0.339 | 6.2 | 4.2 |
| YOLO26 X | 55.7 | 0.303 | 0.158 | 0.412 | 0.239 | 6.1 | 3.5 |

*Table 5. Edge Device - Intel N150, inference with OpenVINO*

| Model | Format | F1-score | Latency (ms) |
|---|---|---|---|
| D-FINE-seg S | FP32 | 0.264 | 431.2 |
| D-FINE-seg S | FP16 | 0.264 | 272.2 |
| D-FINE-seg S | INT8 | 0.243 | 205.0 |
| YOLO26-seg S | INT8 | 0.153 | 113.6 |
| D-FINE S | FP32 | 0.272 | 188.4 |
| D-FINE S | FP16 | 0.271 | 120.8 |
| D-FINE S | INT8 | 0.250 | 76.3 |
| YOLO26 S | INT8 | 0.134 | 63.0 |

On average D-FINE-seg achieved ~**41%** higher mask mAP and ~49% higher box mAP. YOLO26's model size M achieved higher mask mAP, but underperformed on N, S, L, and X sizes.

Finally, we report results for different deployment formats (Table 3) as well as quantized inference on an edge device (Table 5).

## 6. CONCLUSION

D-FINE-seg outperforms YOLO26 in fine-tuning task (reported metrics on TACO dataset) when measuring F1-score with a fixed threshold, showing both better overall accuracy and accuracy-to-latency trade-off. On AP metrics, in segmentation task, the gap between D-FINE-seg and YOLO26 was still meaningful. In detection task D-FINE was a clear winner.

D-FINE-seg is an open-source framework with Apache-2.0 license intended for reproducible training on custom datasets. It supports object detection and instance segmentation tasks and allows to export models and run inference on different hardware.

While the reported results are limited to a specific dataset and fine-tuning setting, they indicate that D-FINE-seg is a strong real-time alternative to YOLO-style instance segmentation. Evaluating on additional datasets and deployment conditions is an important direction for future work.

# 7. REFERENCES

[1] N. Carion, F. Massa, G. Synnaeve, N. Usunier, A. Kirillov and S. Zagoruyko, "End-to-End Object Detection with Transformers", arXiv preprint arXiv:2005.12872, May 2020.

[2] Y. Peng, H. Li, P. Wu, Y. Zhang, X. Sun and F. Wu, "D-FINE: Redefine Regression Task in DETRs as Fine-grained Distribution Refinement", arXiv preprint arXiv:2410.13842, Oct. 2024.

[3] Y. Zhao, W. Lv, S. Xu, J. Wei, G. Wang, Q. Dang, Y. Liu and J. Chen, "DETRs Beat YOLOs on Real-time Object Detection", arXiv preprint arXiv:2304.08069, Apr. 2023.

[4] F. Li, H. Zhang, H. xu, S. Liu, L. Zhang, L. M. Ni, H. Shum, "Mask DINO: Towards A Unified Transformer-based Framework for Object Detection and Segmentation", arXiv preprint arXiv:2206.02777, Jun. 2022.

[5] X. Xu, Y. Jiang, W. Chen, Y. Huang, Y. Zhang and X. Sun, "DAMO-YOLO : A Report on Real-Time Object Detection Design", arXiv preprint arXiv:2211.15444, Nov. 2022.

[6] Z. Ge, S. Liu, F. Wang, Z. Li and J. Sun, "YOLOX: Exceeding YOLO Series in 2021", arXiv preprint arXiv:2107.08430, Jul. 2021.

[7] Ultralytics, "YOLO26," Ultralytics Documentation. Available: https://docs.ultralytics.com/models/yolo26/, [Online], accessed Jan. 2026.

[8] A. Bochkovskiy, C. Wang, H. Liao, "YOLOv4: Optimal Speed and Accuracy of Object Detection", arXiv preprint arXiv:2004.10934, Apr. 2020.

[9] B. Cheng, I. Misra, A. Schwing, A. Kirillov and R. Girdhar, "Masked-attention Mask Transformer for Universal Image Segmentation", arXiv preprint arXiv:2112.01527, Jun. 2022.

[10] H. Zhang, F. Li, S. Liu, L. Zhang, H. Su, J. Zhu, L. Ni and Heung-Yeung Shum "DINO: DETR with Improved DeNoising Anchor Boxes for End-to-End Object Detection", arXiv preprint arXiv:2203.03605, Mar. 2022.

[11] A. Kirillov, E. Mintun, N. Ravi, H. Mao, C. Rolland, L. Gustafson, T. Xiao, S. Whitehead, A. Berg, W. Lo, P. Dollár and R. Girshick, "Segment Anything", arXiv preprint arXiv:2304.02643, Apr. 2023.

[12] ONNX Community, "Open Neural Network Exchange (ONNX)," 2017–present. Available: https://onnx.ai/, [Online], accessed Jan. 2026.

[13] NVIDIA Corporation, "NVIDIA® TensorRT™". Available: https://developer.nvidia.com/tensorrt, [Online], accessed Jan. 2026.

[14] Intel Corporation, "Open-source software toolkit for optimizing and deploying deep learning models". Available: https://github.com/openvinotoolkit/openvino, [Online], accessed Jan. 2026.

[15] H. Zhang, Y. Wang, F. Dayoub and N. Sünderhauf, "VarifocalNet: An IoU-aware Dense Object Detector", arXiv preprint arXiv:2008.13367, Aug. 2020.

[16] T. Lin, P. Goyal, R. Girshick, K. He and P. Dollár, "Focal Loss for Dense Object Detection", arXiv preprint arXiv:1708.02002, Aug. 2017.

[17] H. Rezatofighi, N. Tsoi, J. Gwak, A. Sadeghian, I. Reid and S. Savarese, "Generalized Intersection over Union: A Metric and A Loss for Bounding Box Regression", arXiv preprint arXiv:1902.09630, Feb. 2019.

[18] Github issue - https://github.com/Peterande/D-FINE/issues/108, [Online], accessed Jan. 2026.

[19] P. Proença and P. Simões, "TACO: Trash Annotations in Context for Litter Detection", arXiv preprint arXiv:2003.06975, Mar. 2020.